\documentclass{article}

\usepackage{arxiv}

\usepackage[utf8]{inputenc}
\usepackage[T1]{fontenc}
\usepackage{amsmath}
\usepackage{amsfonts}
\usepackage{amssymb}
\usepackage{booktabs}
\usepackage{graphicx}
\usepackage{microtype}
\usepackage[numbers,sort&compress]{natbib}
\usepackage[hyphens]{url}
\usepackage{wrapfig}
\usepackage{float}
\usepackage{capt-of}
\usepackage{etoolbox}
\usepackage{needspace}
\usepackage{fontawesome5}
\usepackage[hidelinks]{hyperref}
\usepackage{doi}

\newcommand{\Aop}{\mathcal{A}}
\newcommand{\Fop}{\mathcal{F}}

\title{Allocating Recurrent Compute in Looped Language Models}

\author{
  Ruhai Lin\thanks{Equal contribution.} \quad
  Yiyang Guo\footnotemark[1] \quad
  Rui-jie Zhu \quad
  Hao Ye \quad
  Jason Eshraghian\thanks{Corresponding author.}\\
  University of California, Santa Cruz
}

\date{}

\renewcommand{\shorttitle}{Allocating Recurrent Compute in Looped Language Models}

\hypersetup{
  pdftitle={Allocating Recurrent Compute in Looped Language Models},
  pdfauthor={Ruhai Lin, Yiyang Guo, Rui-jie Zhu, Hao Ye, Jason Eshraghian},
  pdfkeywords={looped language models, recurrent depth, Gated DeltaNet, latent reasoning}
}

\makeatletter
\patchcmd{\@maketitle}{\vskip 0.4in \@minus 0.1in \center{\@date}   \vskip 0.2in}{\vskip 12pt}{}{}
\patchcmd{\@maketitle}{\LARGE\sc}{\Large\sc}{}{}
\patchcmd{\@maketitle}{\textsc{\undertitle}\\}{}{}{\PackageError{MixerLoop}{Preprint subtitle patch failed}{Check arxiv.sty.}}
\makeatother
\begin{document}

\raggedbottom

\maketitle

\begin{abstract}
    Looped language models improve reasoning and knowledge manipulation by applying shared computation repeatedly. Existing systems usually repeat an entire layer stack, although a mixer and a dense feed-forward network (FFN) perform different operations and have different costs. We ask a narrower question: what should loop? We view recurrence as iterative refinement of a hidden representation and separate two effects that full-block recurrence couples: repeated context-dependent state updates in the mixer and repeated position-wise transformation in the FFN. This view motivates MixerLoop, which repeats each Gated DeltaNet mixer while applying its dense FFN once. Across matched pretraining experiments, MixerLoop preserves the performance benefits of full-block recurrence while reducing backbone projection compute by 43\%, and consistently outperforms NoLoop. We observe that successive mixer applications continue to reduce next-token cross-entropy. MixerLoop also creates an opportunity during inference because mixer weights are reused across recurrent steps. We implement a custom FPGA accelerator on a KV260 that keeps these weights on chip and overlaps additional mixer computation with the transfer of the larger FFN weights. Using the same hardware and compute datapath, NoLoop and four-step MixerLoop both achieve approximately 152 tokens/s. In memory-bound decoding, MixerLoop therefore turns weight-transfer time into useful recurrent computation with nearly unchanged latency. Together, these results identify the token mixer as an effective recurrence boundary, retaining the benefits of recurrent depth while placing additional computation where it can be efficiently reused by hardware during inference.
\end{abstract}

\begingroup
  \centering\footnotesize
  \setlength{\parskip}{0pt}
  \mbox{\faGithub\enspace\url{https://github.com/ruhai-lin/MixerLoop}}\hspace{1.5em}%
  \mbox{\raisebox{-1.5pt}{\includegraphics[height=9pt]{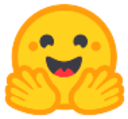}}\enspace
  \url{https://huggingface.co/ruhai-lin/MixerLoop}}\par
\endgroup

\begin{figure}[H]
    \centering
    \hspace*{-5mm}%
    \begin{minipage}{0.85\linewidth}
    \raisebox{2mm}[\height][\depth]{%
        \includegraphics[width=0.45\linewidth]{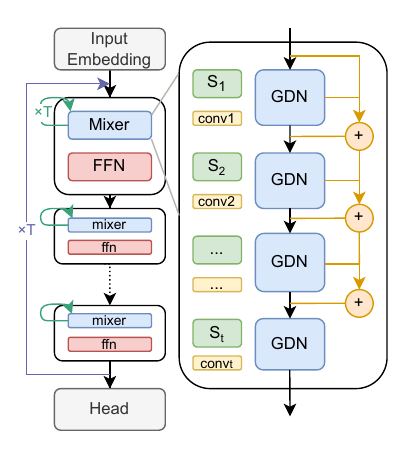}%
    }\hfill%
    \includegraphics[width=0.55\linewidth]{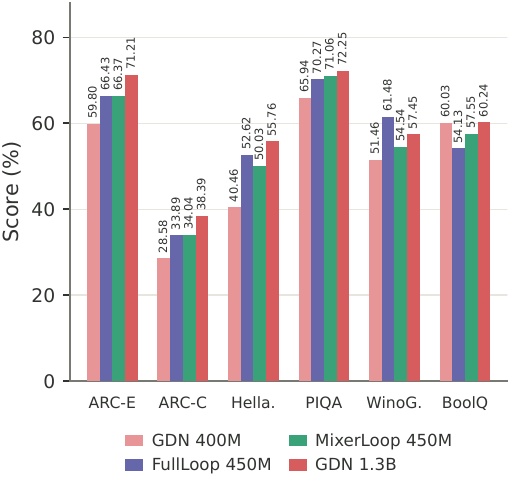}
    \end{minipage}
    \caption{\textbf{(Left)} MixerLoop Architecture; \textbf{(Right)} Evaluation Results and Comparison.} 
    \label{fig:main_fig}
\end{figure}
\clearpage

\section{Introduction}
\label{sec:intro}

Looped language models increase computational depth by repeatedly applying shared parameters, but usually inherit the recurrence boundary from the original feed-forward architecture: the complete block or layer stack. This couples two operators with different roles and costs. The token mixer performs context-dependent interaction and recurrent state updates, while the dense feed-forward network (FFN) applies a large position-wise transformation. Repeating both is therefore only one way to allocate recurrent compute. Recent work has shown that recurrent depth itself is useful~\citep{dehghani2018universal,geiping2026scaling,deng2026lt2}, and SGT further suggests that useful recurrence can be localized within a block~\citep{chen2026sparse}. This motivates a simpler question: what should loop?

We study this question with MixerLoop, which applies the Gated DeltaNet mixer $T$ times in each physical layer followed by a single SwiGLU FFN. Mixer iterations share weights but maintain separate recurrent states across tokens, increasing contextual depth without repeatedly executing the FFN. We compare MixerLoop with matched NoLoop and FullLoop models under the same parameter scale, data, and training budget. MixerLoop achieves the strongest aggregate CORE result at both 13M and 100M, while further analysis shows that successive mixer applications continue to improve next-token prediction. 

This recurrence boundary not only reduces the backbone projection compute by 43\% relative to FullLoop natively, but also creates a corresponding hardware opportunity. Mixer weights are reused immediately across iterations, while the larger FFN weights are needed only once per layer. We implement a complete batch-1 decoder on a KV260 FPGA that caches mixer weights on chip and executes additional mixer iterations while the following FFN weights are transferred. Using the same bitstream and compute datapath, NoLoop and four-step MixerLoop both reach approximately 152 tokens/s. MixerLoop therefore connects recurrent-compute allocation with weight reuse: additional contextual computation can improve the model while, in memory-bound decoding, fitting into time otherwise spent moving weights.


\section{Related work}
\label{sec:related_work}

Looped Transformers partially decouple parameter count from effective computational depth by repeatedly applying shared transformations. Universal Transformer formalized this depth recurrence, and subsequent work showed that shared computation can provide an inductive bias suited to iterative algorithms and use additional depth for latent computation~\citep{dehghani2018universal,yang2024looped,saunshi2025reasoning}. In recent years, this paradigm has extended to large-scale language modeling, but most use a complete Transformer block or layer stack as the recurrent unit~\citep{geiping2026scaling,zhu2025scaling,fu2025think,deng2026lt2,jeddi2026loopformer,lee2026sparse,chen2026loopmoe}. This recurrence boundary also directly determines inference cost. Parameter sharing reduces the number of distinct weights that must be stored, but does not make repeated computation with those weights free. Existing work therefore explores weight locality, dynamic batching, cross-loop parallelism, and intermediate predictions to reorganize recurrent computation, reducing repeated transfers or hiding part of the additional latency in memory-bound decoding~\citep{liu2024mobilellm,bae2025relaxed,wu2025parallel,cho2026loopspec}. These results do show that the actual cost of looped models need not grow linearly with loop count, but still retain the block-level recurrence boundary, so each iteration still requires both mixer and FFN computation.

Prior work also provides grounds for reconsidering this uniform allocation of computation. SGT selectively increases attention depth based on the uncertainty of softmax attention, showing that recurrent computation need not be allocated uniformly across an entire block~\citep{chen2026sparse}. More broadly, studies of functional specialization within Transformers link FFNs to parametric knowledge and distributional associations, while attention more directly supports input-dependent contextual interactions and in-context computation~\citep{geva2021transformer,dai2022knowledge,meng2022locating,meng2022mass,chen2025distributional}. Along similar lines, explicit memory architectures increasingly separate capacity for static or long-tail knowledge from dense neural computation, using sparse access for retrieval~\citep{lample2019large,berges2024memory,pouransari2026pretraining,cheng2026conditional}. Together, these studies motivate assigning different recurrent depths to contextual mixing and parameterized feed-forward transformations.

For state-based mixers, this distinction further affects execution because the recurrence boundary determines how recurrent states and shared weights move through the memory hierarchy. Model-specific FPGA accelerators show that dedicated dataflows organized around a model's dependencies, state accesses, and numerical representations can substantially change execution cost~\citep{zhu2024scalable,wei2025lightmamba,dong2025ssma,wang2025fastmamba}. More directly, a persistent-state GDN accelerator keeps fixed-size recurrent matrix states in on-chip FPGA memory, avoiding per-token transfers of the entire state between external memory and compute units~\citep{gupta2026persistent}. This implementation focuses on the GDN recurrent kernel and does not cover model projections, the FFN, or the output head. Existing hardware work thus shows that state-based mixing can benefit from data-locality optimizations, but how to coordinate this locality with the remaining computation of a complete looped language model remains an open systems problem.

\section{Method}
\label{sec:method}

\subsection{MixerLoop Architecture}
\label{subsec:software_arch}

MixerLoop applies the Gated DeltaNet (GDN) mixer~\citep{yang2025gdn} $T$ times within each physical layer, followed by a single SwiGLU FFN. Across $L$ layers, this gives $LT$ mixer applications and $L$ FFN applications. Table~\ref{tab:architecture} compares the recurrence boundaries, where $\Aop_i$ and $\Fop_i$ denote the mixer and FFN in layer $i$, including RMSNorm pre-normalization and their standard residual connections.

\begin{wraptable}{r}{0.50\textwidth}
\centering
\caption{Recurrence boundaries and mixer/FFN applications per
physical layer. Additional learned residuals are omitted.}
\label{tab:architecture}
\begingroup
\footnotesize
\setlength{\tabcolsep}{3pt}
\renewcommand{\arraystretch}{1.05}
\begin{tabular*}{\linewidth}{@{\extracolsep{\fill}}llc@{}}
\toprule
Model & Computation & $(A,F)$ \\
\midrule
NoLoop
& $\Fop_L\!\circ\!\Aop_L\!\circ\!\cdots$
& $(1,1)$\\
FullLoop
& $(\Fop_L\!\circ\!\Aop_L\!\circ\!\cdots)^T$
& $(T,T)$\\
MixerLoop
& $(\Fop_L\!\circ\!\Aop_L^T)\!\circ\!\cdots$
& $(T,1)$\\
\bottomrule
\end{tabular*}
\endgroup
\end{wraptable}

Iterations within a layer share mixer and normalization weights, but each maintains its own recurrent state and short-convolution history. During decoding, each iteration reads and updates the state left by the same iteration at the previous token. With $T=4$, each physical layer retains four separate state sets: the token representation passes through the four iterations in sequence, while all four state sets persist for the next token. Different physical layers retain distinct weights and states. Each mixer iteration also adds a channel-wise scaled copy of its input to its output~\citep{deng2026lt2}. The scaling coefficients $\rho$ are zero-initialized, specific to each iteration, and shared across physical layers, adding $Td$ parameters for hidden dimension $d$.

\subsection{Hardware Accelerator}
\label{subsec:hardware_arch}

\begin{figure}[!htbp]
    \centering
    \includegraphics[width=\linewidth]{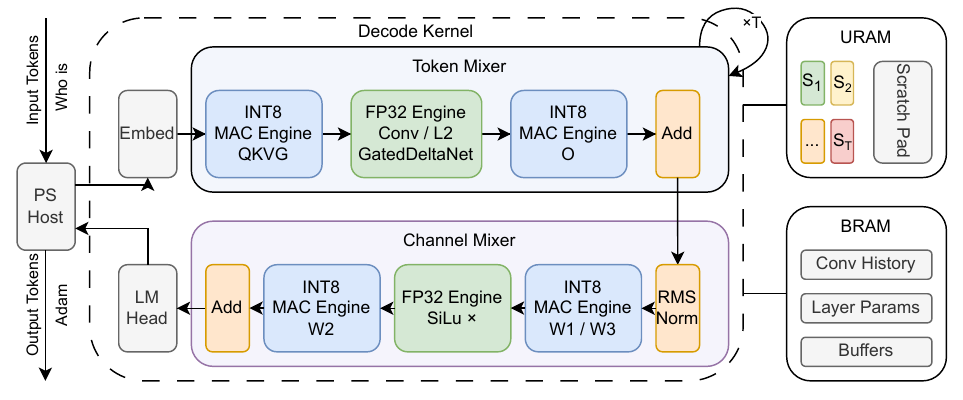}
    \caption{MixerLoop accelerator on KV260. All layers and
    iterations share the projection and recurrent-state datapaths.
    Mixer weights are cached within each layer, while independent
    recurrent states persist across tokens. Additional mixer
    iterations overlap FFN weight prefetch into a streaming
    ring buffer. Computational modules with the same color
    correspond to the same physical block.}
    \label{fig:hardware_arch}
\end{figure}

We next implement a batch-1, single-token decoder for the 13M NoLoop and MixerLoop models on a KV260 FPGA as shown in Figure~\ref{fig:hardware_arch}. The design retains the current layer's mixer weights on chip and overlaps additional applications of $\Aop_i$ with weight transfers for $\Fop_i$. The processing system (PS) handles model loading, tokenization, and kernel invocation, while the programmable logic (PL) executes all model layers and the vocabulary projection, returning the next token and retaining states across invocations.

The decoder uses W8A8 quantization with groups of 32, retaining scales, auxiliary parameters, and persistent states in FP32. A shared 64-MAC INT8 engine executes the mixer Q/K/V/G and output projections, the three FFN projections, and the vocabulary projection. A shared 16-lane FP32 datapath handles recurrent-state updates. Each layer and iteration has separate state and short-convolution storage on chip; matrix states reside in banked URAM.

Within $\Aop_i$, a direct implementation of GDN will traverse the state matrix three times: retrieval forms the delta correction, the correction updates the state, and the updated state produces the output. For one head, let $S^{-}$ and $S^{+}$ denote the old and updated states. The recurrence and output expansion are
\begin{equation}
\begin{aligned}
    \Delta &= \beta\!\left(v-(gS^{-})^{\top}k\right), \qquad S^{+}=gS^{-}+k\Delta^{\top},\\
    o &= (S^{+})^{\top}q = g(S^{-})^{\top}q+(q^{\top}k)\Delta.
\end{aligned}
\label{eq:gdn_output_expansion}
\end{equation}
Here, $q,k,v$ are the query, key, and value, with query scaling absorbed into $q$; $g$ and $\beta$ are the decay and update gates. The expanded output allows \citet{gupta2026persistent} to accumulate retrieval and the old-state output contribution in the same read pass, leaving state update for the second pass.

This fusion places two FP32 vector reductions in the first pass and requires an auxiliary $q^{\top}k$ dot product and output-correction stage. To simplify scheduling on our shared FP32 datapath, we instead accumulate the output during state update:
\begin{equation}
\begin{aligned}
    \text{Pass 1:}\quad
    &\bar{S}=gS^{-},\qquad v_{\mathrm{ret}}=\bar{S}^{\top}k,\\
    &\Delta=\beta(v-v_{\mathrm{ret}}),\\
    \text{Pass 2:}\quad
    &S^{+}=\bar{S}+k\Delta^{\top},\qquad o=(S^{+})^{\top}q.
\end{aligned}
\label{eq:gdn_two_pass}
\end{equation}
The first pass writes the decayed state back to URAM while accumulating retrieval. After computing the delta correction, the second pass reads this state and applies the update, using each updated element immediately for both writeback and output accumulation. Writing and rereading the intermediate state separates the two reductions, leaving one vector accumulation per pass. This also removes the auxiliary dot product and output-correction stage. Interleaved partial sums accommodate FP32 accumulation latency within each pass.

We then organize weight storage around reuse within each physical layer. On-chip weight memory is divided into a mixer cache and an FFN ring buffer. During the first mixer iteration, weights stream from DDR to the projection engine and are simultaneously written into the cache. Subsequent iterations replay these cached weights. FFN weights enter the ring buffer, where consumption immediately releases space for incoming words, allowing the FFN to stream through a buffer smaller than its complete weight set. Projection weights and group scales are organized into 512-bit packed words, transferred through HP0 over a 128-bit AXI interface.

Once the first iteration has fetched the mixer weights, an independent memory process begins FFN prefetch while the remaining mixer iterations execute from the cache. When the final mixer output is ready, the shared projection engine executes $\Fop_i$ using the buffered weight stream, with prefetch continuing as consumed entries are released. Within the FFN, completed SwiGLU activation groups are quantized while later groups are still being computed, preparing inputs for the down-projection. After all physical layers, the same engine streams the vocabulary projection to select the next token.

\subsection{Recurrent Computation}
\label{subsec:recurrent_computation}

\paragraph{Iteration contributions.}
We test the contribution of loop by progressively truncating the mixer iterations of a trained checkpoint. Starting from MixerLoop models trained with four iterations, we retain the first $T\in\{1,2,3,4\}$ mixer applications and their learned residual vectors in each physical layer, followed by its single FFN. Let $\mathcal{L}_T$ denote the mean next-token cross-entropy with $T$ mixer applications per layer. We measure the contribution of increasing the loop budget through \begin{equation} \Delta_T = \mathcal{L}_{T-1}-\mathcal{L}_T, \qquad T=2,3,4, \label{eq:analysis_loop_gain} \end{equation} where positive $\Delta_T$ indicates that the additional iteration per layer improves prediction. We also record top-1 accuracy and the mean probability assigned to the correct next token. To examine variation across layers, we separately change the loop count in one physical layer while retaining four iterations in the others.

\paragraph{Representation geometry.}
To characterize the representations reaching and leaving the FFN, we collect token-aligned activations at selected layers, including the mixer output, normalized FFN input, and block output. These locations let us compare representation geometry across the FFN stage. We measure how broadly token variation is distributed across representation directions using entropy effective rank. For singular values $\sigma_a$ of the centered activation matrix,
\begin{equation}
\pi_a=\frac{\sigma_a}{\sum_b\sigma_b}, \qquad
r_{\mathrm{eff}}=\exp\left(-\sum_a\pi_a\log\pi_a\right).
\label{eq:analysis_effective_rank}
\end{equation}
At the normalized FFN input, we also measure the fraction of squared activation magnitude carried by the largest $10\%$ of coordinates in each token vector, together with mean pairwise cosine similarity within each window. These statistics describe concentration across input coordinates and similarity between token representations, respectively.

\paragraph{FFN response.}
Although the FFN executes only once per layer, it receives the output of all preceding mixer iterations. To examine how additional mixer computation interacts with the once-per-layer FFN, we use two complementary comparisons. First, we vary the mixer count in a single middle or final block while keeping every other block at four iterations and retaining one FFN application in the target block. We measure the resulting change in next-token cross-entropy, top-1 accuracy, and correct-token probability. Second, we compare the effective-rank contrast at the normalized FFN input and residual block output, measuring how the representation difference induced by additional mixer iterations changes across the FFN stage.

\section{Experiments}
\label{sec:experiments}

\subsection{Training and Evaluation}
\label{subsec:train&eval}

We first pre-train NoLoop, MixerLoop, and FullLoop on ClimbMix at the 13M and 100M scales. NoLoop is the single-pass GDN baseline. Within each comparison, the models share the same backbone geometry, tokenizer, corpus, and processed-token budget. The 10B-token runs use the 32K Llama~2 tokenizer, a context length of 1,024, and a global batch of 128 sequences. Both looped variants use a fixed $T=4$ during training. All models are trained with next-token cross-entropy at the final model output, without intermediate-loop supervision.

Training uses the \textit{Flame}~\citep{yang2025flame} framework built on TorchTitan~\citep{liang2025torchtitan}, with chunkwise GDN kernels from Flash Linear Attention. For these 10B-token runs, AdamW uses a peak learning rate of $5\times10^{-4}$, weight decay $0.1$, and 1,000 warmup updates followed by cosine decay to zero over a total of 76,294 updates. We exclude scalar and vector parameters, as well as GDN's \texttt{A\_log} and \texttt{dt\_bias}, from weight decay; all remaining parameters, including the learned residual matrix $\rho$, are decayed. NoLoop and FullLoop retain the FLA and LT2 initialization schemes, respectively; MixerLoop initializes mixer output projections with standard deviation $0.02/\sqrt{2LT}$.

We then evaluate downstream performance on the 22-task CORE V2 suite from DataComp-LM~\citep{li2024datacomp}, using the task definitions and examples in the NanoChat~\citep{karpathy2025nanochat} evaluation bundle. CORE is the unweighted mean of baseline-normalized task accuracies, scaled by 100. Table~\ref{tab:core_main} reports the resulting task and aggregate scores.

Lastly, we train a 450M-scale MixerLoop and evaluate zero-shot benchmarks with LM Evaluation Harness, using the task sets reported by LT2~\citep{deng2026lt2} and Gated DeltaNet~\citep{yang2025gdn}. We report accuracy or length-normalized accuracy according to each task's protocol and note that these results are different from CORE scores.

\subsection{Language Modeling}
\label{subsec:model_performance}

Table~\ref{tab:core_main} organizes the 22 CORE tasks into three
groups: knowledge and commonsense, language and reading comprehension,
and symbolic and analytical reasoning. Group averages summarize the
task profile without changing the original CORE aggregation.

\begin{table}[!t]
\centering
\caption{CORE results by task group. The 13M and 100M models use 10B ClimbMix tokens; the 450M model uses 100B FineWeb-Edu tokens. Scores are baseline-adjusted and scaled by 100.}
\label{tab:core_main}
\begingroup
\footnotesize
\setlength{\tabcolsep}{2pt}
\renewcommand{\arraystretch}{0.9}

\begin{tabular*}{\linewidth}{
    @{\extracolsep{\fill}}
    l
    *{3}{c}
    @{\hspace{8pt}}
    *{3}{c}
    @{\hspace{8pt}}
    c
    @{}
}
\toprule
Task
& \multicolumn{3}{c}{13M}
& \multicolumn{3}{c}{100M}
& 450M \\
\cmidrule(lr){2-4}
\cmidrule(lr){5-7}
\cmidrule(l){8-8}
& NoLoop & MixerLoop & FullLoop
& NoLoop & MixerLoop & FullLoop
& MixerLoop \\
\midrule

\multicolumn{8}{l}{
    \textit{Knowledge and commonsense (8 tasks)}
} \\

Jeopardy
& 0.03 & 0.02 & \textbf{0.06}
& 0.71 & 0.66 & \textbf{1.04}
& 5.15 \\

Wikidata QA
& 4.01 & 3.14 & \textbf{7.19}
& 21.50 & 25.71 & \textbf{28.28}
& 53.69 \\

ARC-Easy
& \textbf{15.51} & 15.21 & 15.17
& 31.76 & 34.12 & \textbf{38.55}
& 51.29 \\

ARC-Challenge
& -3.79 & -5.42 & \textbf{-3.75}
& -0.11 & 1.82 & \textbf{4.55}
& 13.08 \\

COPA
& -11.33 & \textbf{-9.33} & -12.00
& 0.00 & \textbf{2.00} & \textbf{2.00}
& 30.00 \\

CommonsenseQA
& \textbf{-18.03} & -23.42 & -24.66
& -24.43 & \textbf{-1.52} & -30.60
& -25.11 \\

PIQA
& 16.29 & \textbf{16.72} & 16.61
& 31.66 & 33.41 & \textbf{35.58}
& 40.48 \\

OpenBookQA
& 3.56 & 4.18 & \textbf{4.27}
& 9.07 & 8.80 & \textbf{9.87}
& 16.53 \\

\addlinespace[1pt]
\textit{Group mean}
& \textbf{0.78} & 0.14 & 0.36
& 8.77 & \textbf{13.12} & 11.16
& 23.14 \\

\midrule

\multicolumn{8}{l}{
    \textit{Language and reading comprehension (9 tasks)}
} \\

HellaSwag (0-shot)
& \textbf{3.49} & 3.35 & 2.97
& 13.28 & 13.38 & \textbf{16.72}
& 32.38 \\

HellaSwag
& 2.89 & \textbf{2.93} & 2.35
& 12.59 & 12.67 & \textbf{16.35}
& 31.39 \\

LAMBADA
& 10.58 & \textbf{11.74} & 11.68
& 25.36 & 24.63 & \textbf{26.35}
& 39.86 \\

Winograd
& \textbf{7.20} & 3.05 & 5.01
& \textbf{17.22} & 11.36 & \textbf{17.22}
& 38.46 \\

WinoGrande
& 2.76 & 2.03 & \textbf{3.18}
& 2.76 & 0.08 & \textbf{2.92}
& 9.71 \\

Language ID
& 0.17 & \textbf{0.77} & 0.28
& -0.35 & \textbf{1.00} & 0.25
& 0.29 \\

SQuAD
& \textbf{1.38} & 0.85 & 1.15
& \textbf{7.08} & 4.71 & 4.97
& 14.35 \\

CoQA
& \textbf{4.41} & 3.15 & 3.61
& 11.34 & 11.66 & \textbf{12.39}
& 18.05 \\

BoolQ
& -40.89 & -36.17 & \textbf{-35.44}
& -18.78 & -5.91 & \textbf{-5.42}
& -32.95 \\

\addlinespace[1pt]
\textit{Group mean}
& -0.89 & -0.92 & \textbf{-0.58}
& 7.83 & 8.18 & \textbf{10.19}
& 16.84 \\

\midrule

\multicolumn{8}{l}{
    \textit{Symbolic and analytical reasoning (5 tasks)}
} \\

Dyck
& 0.13 & \textbf{0.83} & 0.57
& 12.80 & \textbf{17.20} & 9.30
& 3.20 \\

LSAT-AR
& -5.12 & \textbf{-0.48} & -1.64
& -1.45 & \textbf{7.25} & 4.93
& 2.03 \\

CS Algorithms
& 23.94 & \textbf{30.66} & 22.17
& 35.15 & 35.15 & \textbf{35.23}
& 34.47 \\

Operators
& 13.49 & \textbf{13.97} & \textbf{13.97}
& \textbf{9.52} & 9.05 & 9.05
& 14.29 \\

Repeat-Copy
& \textbf{0.00} & \textbf{0.00} & \textbf{0.00}
& \textbf{0.00} & \textbf{0.00} & \textbf{0.00}
& 6.25 \\

\addlinespace[1pt]
\textit{Group mean}
& 6.49 & \textbf{9.00} & 7.01
& 11.21 & \textbf{13.73} & 11.70
& 12.05 \\

\midrule
\textbf{CORE}
& 1.39 & \textbf{1.72} & 1.49
& 8.94 & \textbf{11.24} & 10.89
& 18.04 \\
\bottomrule
\end{tabular*}
\endgroup
\end{table}

\noindent\begin{minipage}[t]{0.64\linewidth}
\vspace{0pt}
\setlength{\parskip}{5.5pt}
MixerLoop achieves the highest aggregate CORE score at both 13M and 100M (Table~\ref{tab:core_main}). The two recurrence boundaries favor different task groups: MixerLoop leads in symbolic and analytical reasoning at both scales, whereas FullLoop leads in language and reading comprehension.

In the 13M training run, MixerLoop benefits from recurrence earlier than FullLoop (Figure~\ref{fig:13m_training_loss}). It maintains lower loss through the early and middle stages, while FullLoop initially trails NoLoop. The two looped models finish at nearly the same loss, both below NoLoop.

The 450M MixerLoop, trained on 100B FineWeb-Edu tokens, remains close to LT2's FullLoop-GDN in the larger-scale zero-shot comparison (Table~\ref{tab:external_eval}). It closely matches FullLoop on ARC and scores higher on PIQA and BoolQ, with its deficits concentrated in HellaSwag and WinoGrande. MixerLoop also comes within four percentage points of the 1.3B GDN's average score with roughly a third as many parameters.
\end{minipage}\hfill%
\begin{minipage}[t]{0.32\linewidth}
\vspace{-16pt}
\centering
\includegraphics[width=\linewidth]{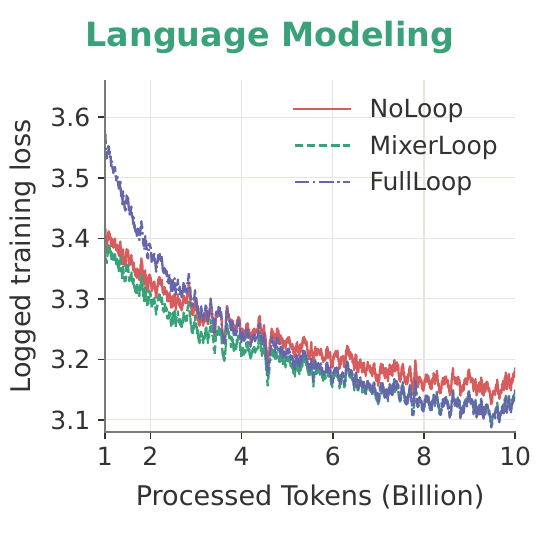}
\captionof{figure}{\footnotesize 13M ClimbMix training (seed 1337), shown from 1B tokens.}
\label{fig:13m_training_loss}
\end{minipage}

\begin{table}[!htbp]
\centering
\caption{Six-task zero-shot results (\%). All models are trained on FineWeb-Edu.}
\label{tab:external_eval}
\begingroup
\footnotesize
\setlength{\tabcolsep}{1.5pt}
\renewcommand{\arraystretch}{1.05}

\begin{tabular*}{\linewidth}{@{\extracolsep{\fill}}lcc*{7}{c}@{}}
\toprule
Model & Scale & Tokens & ARC-E & ARC-C & Hella. & PIQA & WinoG. & BoolQ & Avg. \\
\midrule

NoLoop~\citep{behrouz2025titans} & 400M & 15B
& 59.80 & 28.58 & 40.46 & 65.94 & 51.46 & 60.03 & 51.05 \\

FullLoop~\citep{deng2026lt2} & 450M & 100B
& 66.43 & 33.89 & 52.62 & 70.27 & 61.48 & 54.13 & 56.47 \\

MixerLoop & 450M & 100B
& 66.37 & 34.04 & 50.03
& 71.06 & 54.54 & 57.55
& 55.60 \\

NoLoop~\citep{yang2025gdn} & 1.3B & 100B
& 71.21 & 38.39 & 55.76 & 72.25 & 57.45 & 60.24 & 59.22 \\

\bottomrule
\end{tabular*}

\vspace{2pt}
\raggedright
\scriptsize
LT2 reports the model as 0.6B; the scale shown here follows its code configuration.
The 400M baseline follows the Gated DeltaNet recipe and is trained on 15B FineWeb-Edu tokens.
\endgroup
\end{table}

\Needspace{16\baselineskip}
\subsection{Inference Efficiency}
\label{sec:inference_efficiency}

\setlength{\intextsep}{-8pt}
\begin{wrapfigure}{r}{0.29\textwidth}
    \centering
    \setlength{\abovecaptionskip}{2pt}
    \includegraphics[width=0.95\linewidth]{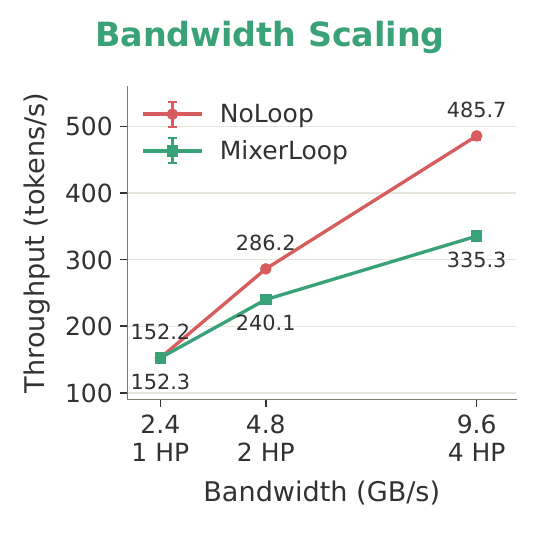}
    \caption{\footnotesize KV260 decode throughput as available DDR bandwidth increases.}
    \label{fig:bandwidth_scaling}
\end{wrapfigure}

MixerLoop preserves the mixer depth of FullLoop while executing the FFN only once per physical layer. Let $C_A$ and $C_F$ denote the projection costs of one mixer and one FFN, respectively. With $T$ iterations, MixerLoop requires $TC_A+C_F$ projection compute per layer, whereas FullLoop requires $T(C_A+C_F)$. Thus, for $r=C_F/C_A$, the reduction is $(T-1)r/[T(1+r)]$. Using $T=4$ and $r\approx1.37$ gives approximately $43.3\%$. Actual decode latency also depends on weight movement and scheduling.

On KV260 at 150 MHz, quantized NoLoop and MixerLoop use the same bitstream and compute datapath, differing only in model weights/checkpoints. We perform eight paired runs per model using the same input prefix, with 128 token-by-token decode calls per run. After discarding the first paired run as warm-up, we report the median of the remaining seven. NoLoop and MixerLoop achieve 152.234 and 152.307 tokens/s, respectively. Figure~\ref{fig:bandwidth_scaling} shows that this overlap diminishes as DDR bandwidth increases and the FFN-transfer window shortens, motivating the deployment condition below. We further compare the implementation with prior language-model accelerators on the same KV260 platform in Table~\ref{tab:fpga_comparison}.

\WFclear
\setlength{\intextsep}{12pt}

\begin{table}[!htbp]
\centering
\caption{Language-model accelerator results on KV260.}
\label{tab:fpga_comparison}
\begingroup
\footnotesize
\setlength{\tabcolsep}{1.5pt}
\renewcommand{\arraystretch}{1.05}
\begin{tabular*}{\linewidth}{@{\extracolsep{\fill}}ll*{7}{c}@{}}
\toprule
Work & Model & MHz & tok/s & LUT & FF & DSP & BRAM36 & URAM \\
\midrule
\citet{li2025pushing} & LLaMA2-7B-Q4 & 300 & 4.9 & 78K & 105K & 291 & 36.5 & 10 \\
Hummingbird~\citep{li2025hummingbird} & LLaMA3-8B-Q4 & 300 & 4.8 & 26K & 35K & 179 & 59.0 & 18 \\
TeLLMe v2~\citep{qiao2025tellmev2} & BitNet-0.73B-Q1.58 & 250 & 25.0 & 98K & 137K & 610 & 98.5 & 60 \\
MixerLoop & MixerLoop-13M-Q8 & 150 & 152.3 & 79K & 107K & 453 & 81.0 & 64 \\
\bottomrule
\end{tabular*}
\endgroup
\end{table}

The verified overlap suggests a design rule for MixerLoop: additional iterations must fit both the time left by FFN weight transfer and the capacity available for prefetching. For decoding with batch size $b$, let $t_{\Aop}(b)$ and $t_{\Fop}(b)$ denote the compute service times of one cached mixer application and one FFN, excluding external-weight stalls. Let $W_{\Fop}$ be the layer's FFN weight-transfer volume and $B$ the effective external-memory bandwidth. Of the available on-chip capacity $M$, $M_{\mathrm{res}}(b,T)$ is reserved for the mixer cache, persistent states, and baseline workspace. Under a constant-rate streaming approximation, additional iterations can be hidden when
\begin{equation}
\underbrace{(T-1)t_{\Aop}(b)}_{\text{extra mixer time}}
\leq
\min\!\left\{
\underbrace{\frac{W_{\Fop}}{B}-t_{\Fop}(b)}_{\text{weight-transfer slack}},
\quad
\underbrace{\frac{M-M_{\mathrm{res}}(b,T)}{B}}_{\text{prefetch-buffer slack}}
\right\}.
\label{eq:mixerloop_deployment}
\end{equation}
This condition links recurrent depth, model geometry, and hardware capabilities. On fixed hardware, it guides the choice of a resident mixer and the number of iterations that fit the available slack; for a fixed model, it determines the required compute and buffering. More on-chip memory relaxes the capacity constraint, whereas higher bandwidth shortens the transfer window available for hiding recurrence. Appendix~\ref{app:hardware_deployment} provides the detailed derivation and a KV260 instantiation.

\subsection{Leveraging Recurrence}
\label{subsec:recurrent_results}

\begin{figure}[H]
    \centering
    \includegraphics[width=0.9\linewidth]{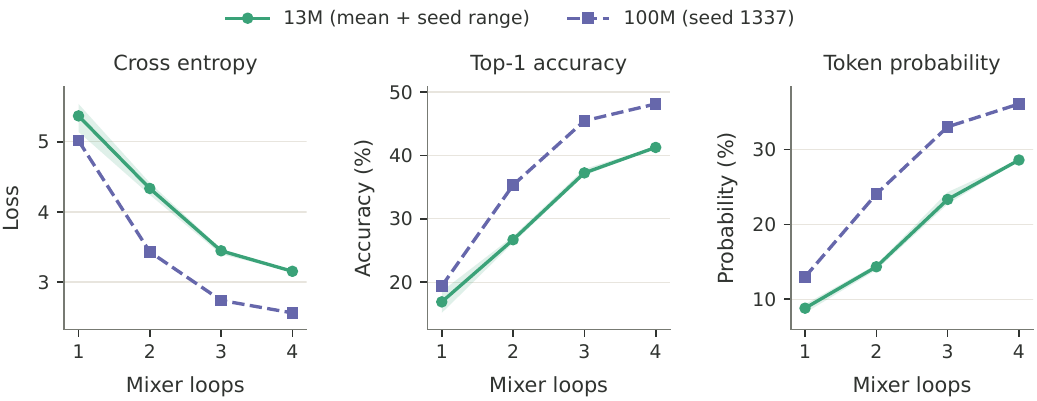}
    \caption{Fixed-checkpoint loop-count sweeps for 13M and 100M ClimbMix MixerLoop. The 13M curve and shading show the mean and range across three seeds; the 100M curve uses seed 1337.}
    \label{fig:loop_quality}
\end{figure}

On 205 validation windows, increasing the loop budget from one to four improves next-token prediction across the three 13M ClimbMix checkpoints and the 100M checkpoint with seed 1337 (Figure~\ref{fig:loop_quality}). Cross-entropy decreases while top-1 accuracy and the mean probability assigned to the correct next token increase. For the 100M model, cross-entropy falls from 5.016 to 2.558, with successive reductions of 1.586, 0.693, and 0.179 nats per token. The gains diminish with recurrent depth, but the fourth iteration still contributes to prediction.

\begin{figure}[H]
    \centering
    \includegraphics[width=0.9\linewidth]{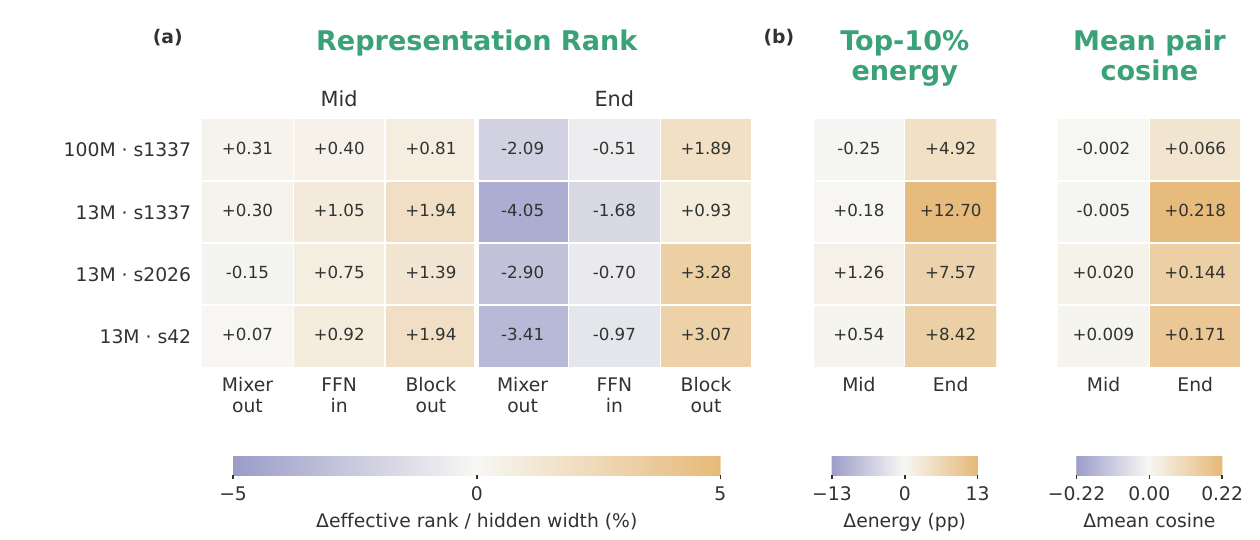}
    \caption{Within-checkpoint differences ($T=4$ minus $T=1$) for 13M and 100M ClimbMix MixerLoop. (a) Centered effective-rank change as a percentage of hidden width. (b) Changes in FFN-input top-$10\%$ coordinate-energy share (percentage points) and mean within-window pairwise cosine. Mid and End denote the middle and final layers.}
    \label{fig:rank_ffn_input}
\end{figure}

The block-local intervention shows that this effect remains predictive when the FFN itself is not repeated. Increasing only the middle block from one to four mixer applications reduces cross-entropy by 0.069--0.205 nats per token, while the same intervention at the final block reduces it by 0.330--1.318. The final-block gain is larger in all four checkpoints, and both top-1 accuracy and correct-token probability improve in all eight checkpoint--layer conditions (Appendix~\ref{app:recurrent_diagnostics}).

The final FFN also transforms the representation contrast induced by recurrence. Across the four checkpoints, the $T=4$ minus $T=1$ effective-rank contrast changes from $-1.68$--$-0.51\%$ of hidden width at the normalized FFN input to $+0.93$--$+3.28\%$ at the residual block output, a shift of $2.40$--$4.04$ percentage points across the FFN stage. Thus, the representation change introduced by additional mixer iterations is carried through and transformed by the single FFN stage before prediction.

Additional mixer computation therefore improves prediction while changing the input presented to the once-per-layer FFN. Its benefit varies with recurrent depth and layer position, while representation concentration alone does not establish how the FFN contributes to that benefit. For recurrence allocation, the marginal prediction gains provide a direct basis for weighing additional iterations against the execution and storage constraints in Section~\ref{sec:inference_efficiency}.

\section{Conclusion}
\label{conclusion}
We presented MixerLoop that assigns additional iterations to the Gated DeltaNet mixer while keeping one FFN application per physical layer. Across matched language-modeling experiments, this boundary preserves the benefits of recurrence while avoiding repeated dense feed-forward computation, and successive mixer applications continue to improve prediction. The same structure also enables weight reuse during decoding: cached mixer iterations can execute while the following FFN weights are transferred, allowing four-step MixerLoop to match NoLoop throughput on KV260 at the evaluated operating point. More broadly, the useful recurrent depth of a model is jointly determined by its predictive gain, operator balance, memory bandwidth, and on-chip state capacity.

\newpage

\begingroup
\small
\bibliography{references}
\bibliographystyle{unsrtnat}
\endgroup

\clearpage
\appendix

\section{Training Details}
\label{app:training}

Table~\ref{tab:model_configs} lists the model geometries. All models use tied input and output embeddings, short convolutions of size four, and value expansion two. MixerLoop and FullLoop use $T=4$ and add a shared residual matrix with $4d$ parameters. The 13M and 100M comparisons include all three architectures; the 450M experiment trains MixerLoop.

\begin{table}[!ht]
\centering
\caption{Model configurations. Parameter counts include tied embeddings and the shared residual matrix. NoLoop has $4d$ fewer parameters at the same geometry.}
\label{tab:model_configs}
\begingroup
\footnotesize
\setlength{\tabcolsep}{3pt}
\begin{tabular*}{\linewidth}{@{\extracolsep{\fill}}lrrrrrrrrr@{}}
\toprule
Scale & $d$ & $L$ & $N_h$ & $d_k$ & $d_v$ & $f$ & $V$ & Context & Parameters \\
\midrule
13M & 256 & 5 & 6 & 32 & 64 & 704 & 32,000 & 1,024 & 12,896,380 \\
100M & 768 & 9 & 9 & 64 & 128 & 2,112 & 32,000 & 1,024 & 100,447,266 \\
450M & 1,024 & 21 & 8 & 96 & 192 & 2,816 & 128,256 & 4,096 & 445,775,120 \\
\bottomrule
\end{tabular*}
\endgroup
\end{table}

Here $d$ is the hidden width, $L$ the number of physical layers, $N_h$ the number of heads, $d_k$ and $d_v$ the key and value head widths, $f$ the FFN intermediate width, and $V$ the vocabulary size. The 13M and 100M models use the Llama~2 tokenizer; the 450M model uses the Llama~3 tokenizer.

\begin{table}[!ht]
\centering
\caption{Training configurations. Token counts include every processed input token.}
\label{tab:training_configs}
\begingroup
\footnotesize
\setlength{\tabcolsep}{5pt}
\renewcommand{\arraystretch}{1.08}
\begin{tabular*}{\linewidth}{@{\extracolsep{\fill}}lll@{}}
\toprule
Setting & 13M / 100M & 450M \\
\midrule
Corpus & ClimbMix & FineWeb-Edu, sample-100BT \\
Framework & FLAME / TorchTitan & LitGPT / Lightning Fabric \\
Processed tokens & 10,000,007,168 & 100,270,080,000 \\
Optimizer updates & 76,294 & 255,000 \\
Global batch, sequences & 128 & 96 \\
Tokens per update & 131,072 & 393,216 \\
Optimizer & Fused AdamW & Fused AdamW \\
Peak learning rate & $5\times10^{-4}$ & $3\times10^{-4}$ \\
Linear warmup, updates & 1,000 & 10,000 \\
Decay & Cosine & Cosine \\
Minimum learning rate & $0$ & $3\times10^{-10}$ \\
AdamW $(\beta_1,\beta_2)$ & $(0.9,0.95)$ & $(0.9,0.95)$ \\
AdamW $\epsilon$ & $10^{-8}$ & $10^{-8}$ \\
Weight decay & $0.1$ & $0.1$ \\
Gradient clipping, norm & $1.0$ & $1.0$ \\
Precision & BF16 AMP, FP32 parameters & BF16 mixed precision \\
Run seeds & 42, 1337, 2026 / 1337 & 1337 \\
\bottomrule
\end{tabular*}
\endgroup
\end{table}

The 13M results average seeds 42, 1337, and 2026; the 100M result uses seed 1337. The 450M seed is recorded in the run label. Its checkpoint contains 255,000 completed updates, and the supplied training configuration uses 96 sequences of 4,096 tokens per update. Microbatch and gradient accumulation vary with the device configuration while preserving the global batch. The 450M run uses fully sharded data parallelism.

AdamW excludes parameters with fewer than two dimensions, \texttt{A\_log}, and \texttt{dt\_bias} from weight decay. Matrix parameters, including the shared residual matrix, receive weight decay. NoLoop retains the FLA initialization. MixerLoop uses normal initialization with standard deviation 0.02 and scales each mixer output projection to $0.02/\sqrt{2LT}$; its shared residual matrix starts at zero. FullLoop follows LT2: linear and embedding weights use a truncated normal with pre-truncation standard deviation $d^{-1/2}$, bounded at three standard deviations, without depth scaling. Its convolutions retain native initialization and its shared residual matrix starts at zero.

ClimbMix is tokenized from the first 170 shards of the shuffled corpus, with shard 6542 reserved for validation. Documents begin with a BOS token. The training loader shuffles shards and fixed-length windows using the run seed and preserves its state across checkpoint resumes. CORE uses the NanoChat evaluation bundle and applies the CORE V2 baseline corrections for CommonsenseQA, LSAT-AR, and language identification. For task accuracy $a_k$ and baseline $c_k$, the reported task score is $100(a_k-c_k)/(1-c_k)$; CORE averages the 22 task scores with equal weight.

\section{Hardware Deployment Analysis}
\label{app:hardware_deployment}

\subsection{Overlap Condition}

For a decode batch of $b$ sequences, the first mixer application streams weights into the current layer's cache. The remaining $T-1$ applications replay this cache while the FFN weights arrive. Let $t_{\Aop}(b)$ include all operations of a cached mixer application, and let $t_{\Fop}(b)$ be the FFN service time excluding external-weight stalls. With FFN transfer volume $W_{\Fop}$ and effective bandwidth $B$,
\begin{equation}
t_{\mathrm{mem}}=\frac{W_{\Fop}}{B},
\qquad
\tau_{\mathrm{extra}}=(T-1)t_{\Aop}(b).
\label{eq:deployment_times}
\end{equation}
We use a constant-rate streaming approximation with independent transfer and computation, continuous FFN weight consumption, and no startup or drain overhead. If $t_{\mathrm{mem}}\geq t_{\Fop}(b)$, recurrence fits within the transfer window when
\begin{equation}
\tau_{\mathrm{extra}}+t_{\Fop}(b)\leq t_{\mathrm{mem}}.
\label{eq:deployment_time_constraint}
\end{equation}

The corresponding storage constraint reserves the current layer's mixer cache, workspace, and all persistent states:
\begin{equation}
M_{\mathrm{res}}(b,T)=W_{\Aop}+M_0+bT M_s,
\label{eq:deployment_reserved_capacity}
\end{equation}
where $W_{\Aop}$ is the mixer-cache footprint, $M_0$ includes workspace and baseline buffer occupancy, and $M_s$ covers one sequence and one iteration index across all physical layers. For homogeneous GDN layers, the matrix-state contribution is $sLN_hd_kd_v$, with $s$ bytes per state element; short-convolution histories are included as well. Capacities include packing and scale overheads.

Before the FFN begins consuming weights, prefetch accumulates an additional $B\tau_{\mathrm{extra}}$ bytes. For available on-chip capacity $M$, uninterrupted prefetch requires
\begin{equation}
B\tau_{\mathrm{extra}}\leq M-M_{\mathrm{res}}(b,T).
\label{eq:deployment_capacity_constraint}
\end{equation}
In this streaming regime, the FFN's stall-free consumption rate is at least the arrival rate, so the additional backlog does not increase after FFN execution starts. Combining the time and capacity constraints gives Equation~\ref{eq:mixerloop_deployment}. These are sufficient conditions within the streaming approximation, rather than measured transition boundaries.

A projection-dominated approximation relates this budget to model geometry. Let $C_A$ and $C_F$ be the per-sequence projection FLOPs, $P$ the sustained projection throughput in MAC/s, and $q_w$ the transferred bytes per matrix weight including scales. Then $t_{\Aop}(b)\approx bC_A/(2P)$, $t_{\Fop}(b)\approx bC_F/(2P)$, and $W_{\Fop}\approx q_wC_F/2$. With $r=C_F/C_A$ and $\kappa=P/B$, the time constraint becomes
\begin{equation}
b\left(1+\frac{T-1}{r}\right)\leq q_w\kappa.
\label{eq:deployment_balance}
\end{equation}
Batch size, recurrent depth, and the FFN-to-mixer ratio determine the compute demand during transfer. The throughput, bandwidth, and storage capacity must refer to the same executable dataflow and memory hierarchy.

\subsection{KV260 Instantiation}

The 13M implementation caches 6,912 mixer words and transfers 9,504 FFN words per layer, with 64 bytes per word. A cached mixer application takes 8,920 cycles. At 150 MHz, the ideal HP0 transfer rate is one word every four cycles. Using the FFN's 9,504-cycle Q8 service as a first-order estimate gives Table~\ref{tab:kv260_overlap}.

\begin{table}[!ht]
\centering
\caption{Per-layer overlap budget at the single-port KV260 operating point.}
\label{tab:kv260_overlap}
\begin{tabular}{@{}lr@{}}
\toprule
Quantity & Time ($\mu$s) \\
\midrule
FFN weight transfer & 253.44 \\
FFN Q8 service & 63.36 \\
Available transfer slack & 190.08 \\
One cached mixer application & 59.47 \\
Three additional mixer applications & 178.40 \\
\bottomrule
\end{tabular}
\end{table}

The time budget permits three additional mixer applications, giving $T_{\max,\mathrm{time}}=1+\lfloor190.08/59.47\rfloor=4$. During these applications, prefetch accumulates 418.1 KiB. Including baseline occupancy, the peak is 446.6 KiB, or 7,145 words, below the 9,472-word ring capacity of 592 KiB. The budget omits complete FFN startup, dependency, and drain overheads; Figure~\ref{fig:bandwidth_scaling} separately reports the measured effect of increasing memory bandwidth.

\section{Additional Recurrent Diagnostics}
\label{app:recurrent_diagnostics}

\begin{figure}[!ht]
    \centering
    \includegraphics[width=\linewidth]{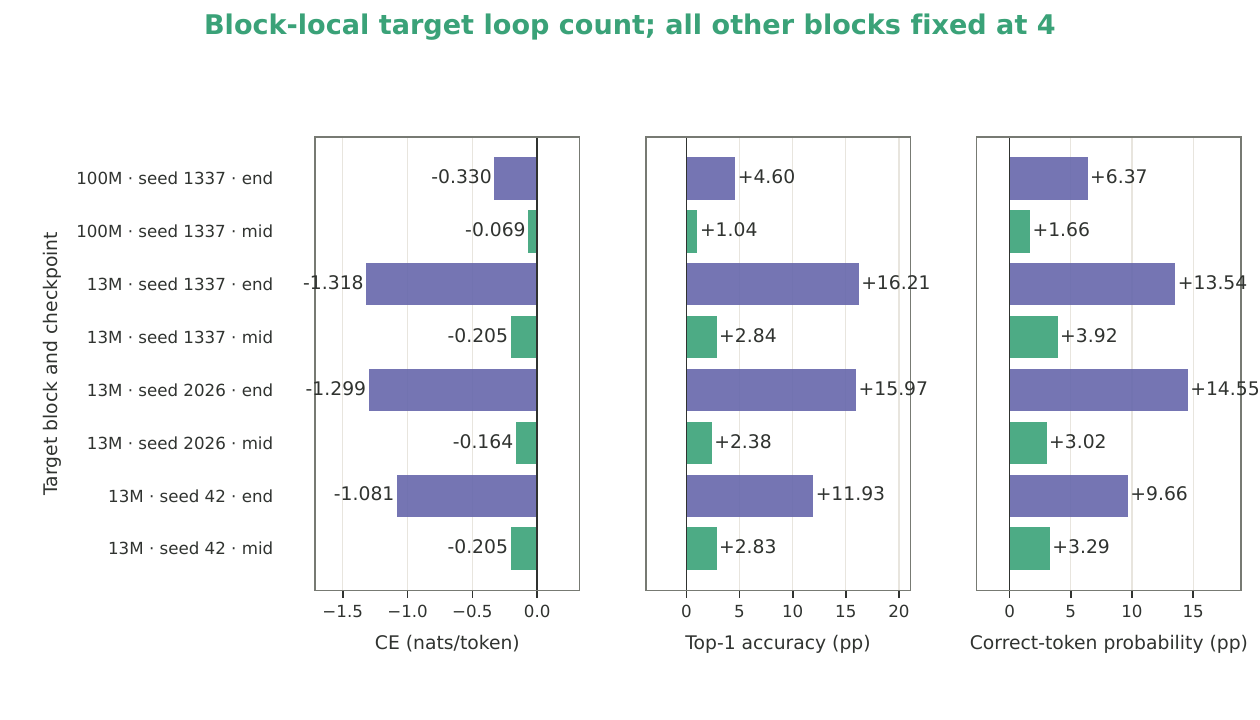}
    \caption{Block-local loop-count sweep on ClimbMix checkpoints. Only the indicated middle or final block changes from one to four mixer calls. Cross-entropy decreases and both prediction metrics increase in all eight checkpoint--layer conditions.}
    \label{fig:block_local_loop}
\end{figure}

The block-local sweep uses three 13M ClimbMix checkpoints with seeds 42, 1337, and 2026, and the 100M ClimbMix checkpoint with seed 1337. It changes the mixer-call count in one middle or final physical block while retaining four calls in every other block and a single FFN per block. Each condition uses the same 205 validation windows. Figure~\ref{fig:block_local_loop} reports four-minus-one differences in the mean of 13 validation-batch scores; cross-entropy is in nats per token, and accuracy and correct-token probability differences are in percentage points.

The representation-rank diagnostics pool fit and validation captures, using 490,496 token rows from 1,024 windows for each condition. We center the activation matrices and use the singular-value entropy definition in Equation~\ref{eq:analysis_effective_rank}. Figure~\ref{fig:rank_ffn_input} reports $100(r_{\mathrm{eff}}^{(4)}-r_{\mathrm{eff}}^{(1)})/d$, where the superscript denotes the loop count and $d$ is 256 for 13M and 768 for 100M.

For the FFN-input diagnostics in Figure~\ref{fig:rank_ffn_input}, the top-$10\%$ coordinate-energy share is the squared magnitude in the largest $\lceil0.1d\rceil$ coordinates divided by the total squared magnitude of each token vector. Pairwise cosine measures within-window similarity.

\end{document}